\documentclass[letterpaper, 10 pt, conference]{ieeeconf}  
\IEEEoverridecommandlockouts                              
\usepackage{xcolor}
\usepackage{soul}
\usepackage{booktabs}
\usepackage{graphicx}
\usepackage{amsmath} 
\graphicspath{ {figures/} }

\begin{document}
\title{\LARGE \bf

Thumb Stabilization and Assistance in a\\Robotic Hand Orthosis for Post-Stroke Hemiparesis 
}

\author{Ava Chen$^{1}$, Lauren Winterbottom$^{2}$, Sangwoo Park$^{1}$,  Jingxi Xu$^{3}$, Dawn Nilsen$^{2,4}$,\\ Joel Stein$^{2,4}$, and Matei Ciocarlie$^{1,4}$%
\thanks{This work was supported in part by the National Institute of Neurological Disorders and Stroke under grant R01NS115652}
\thanks{$^{1}$Department of Mechanical Engineering, Columbia University, New York, NY 10027, USA. {\texttt{\footnotesize \{ava.chen, sp3287, matei.ciocarlie\}@columbia.edu}}}%
\thanks{$^{2}$Department of Rehabilitation and Regenerative Medicine, Columbia University, New York, NY 10032, USA. {\texttt{\footnotesize\{lbw2136, dmn12, js1165\}@cumc.columbia.edu}}}%
\thanks{$^{3}$Department of Computer Science, Columbia University, New York, NY 10027, USA. {\texttt{\footnotesize jxu@cs.columbia.edu}}}%
\thanks{$^{4}$Co-Principal Investigators}
}

\maketitle
\thispagestyle{empty}
\pagestyle{empty}

\begin{abstract}
We propose a dual-cable method of stabilizing the thumb in the context of a hand orthosis designed for individuals with upper extremity hemiparesis after stroke. This cable network adds opposition\slash reposition capabilities to the thumb, and increases the likelihood of forming a hand pose that can successfully manipulate objects. In addition to a \textit{passive-thumb version} (where both cables are of fixed length), our approach also allows for a single-actuator \textit{active-thumb version} (where the extension cable is actuated while the abductor remains passive), which allows a range of motion intended to facilitate creating and maintaining grasps. We performed experiments with five chronic stroke survivors consisting of unimanual resistive-pull tasks and bimanual twisting tasks with simulated real-world objects; these explored the effects of thumb assistance on grasp stability and functional range of motion. Our results show that both active- and passive-thumb versions achieved similar performance in terms of improving grasp force generation over a no-device baseline, but active thumb stabilization enabled users to maintain grasps for longer durations.
\end{abstract}

\section{Introduction}

Hand impairment due to loss of volitional control of the digits is a common contributor to chronic disability following a stroke \cite{kamper2006}. Inability to achieve finger and thumb extension is strongly associated with severity of motor impairment, but is the slowest and least likely movement to recover even following targeted rehabilitation techniques \cite{lang2009}. Recent developments in wearable robotics to assist hand-opening show promise in providing functional support for activities of daily living (ADLs) and encouraging use of the impaired limb outside the clinic \cite{chen2017,gassar2020,lee2021}; we have established that such support can be accomplished using lightweight, underactuated designs \cite{park2020,park2019,park2018}. But the vast majority of robotic orthoses, including our previous work, focus on finger actuation while splinting the thumb in a position of general opposition. Devices that actively assist thumb motion have typically actuated a subset of degrees of freedom contributing to overall opposition\slash reposition \cite{sandison2020,ge2020,gerez2020}, but these devices do not consider whether their methods add value to functional manipulation beyond that provided by a passive thumb splint \cite{se2018}.

\begin{figure}[t]
    \centering
    \includegraphics[width=.96\columnwidth]{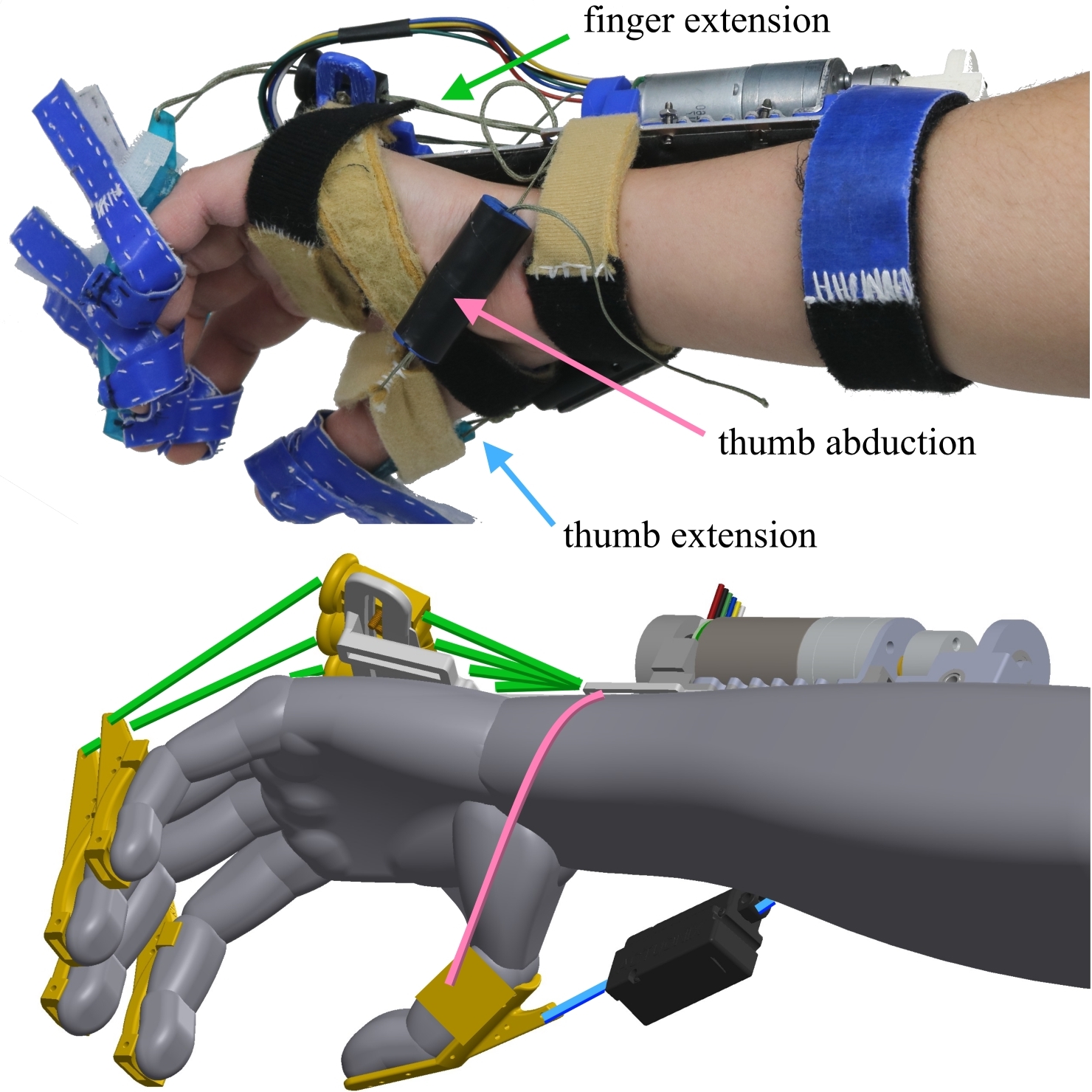}
    \caption{The hand orthosis assists hand-opening for stroke survivors who are unable to actively extend their digits. The active-thumb version has a linear actuator mounted in-line with the thumb's extension cable (blue). A fixed-length cable replaces the linear actuator in the passive version. Thumb \mbox{ab\slash adduction} is controlled by an adjustable passive cable (pink) that wraps around the hand to terminate on the splint at the wrist. A padded tube prevents the cable from digging into the skin as it bends around the base of the hand. A motorized winch pulls a separate tendon network (green) to extend the fingers.}
    \label{fig:myhand}
    \vspace{-0.3cm}
\end{figure}

Thumb placement, and its effects on hand configuration, plays a vital role in how humans interact with the environment around them \cite{parry2019}. One clear benefit for splinting the thumb for stroke survivors is simply to move the thumb out of the way and into a position where it can support oppositional grasping at all, since spastic flexion normally keeps the thumb curled into the palm. Intuitively, expanding the thumb's active range of motion generally increases the hand's ability to open and grasp a greater range of object sizes. However, involuntary synergies and spasticity after stroke often cause asymmetric, abnormal motor coupling between fingers and thumb such that overall hand aperture can decrease with applied extension to a digit \cite{kamper2014}; we  have informally observed such thumb-flexion reflexes in multiple subjects. Robotic interventions intended to support ADLs for a stroke population must overcome additional challenges to maintain grasp stability against external and stroke-derived perturbations in order to perform better than baseline compensatory techniques.

In this study, we propose a dual-cable method to assist thumb placement: one cable aligned with the thumb's flexion\slash extension degree of freedom (DoF) and the other with \mbox{ab\slash adduction}. Both cables jointly contribute to thumb extension and abduction; for simplicity we refer to each tendon by its primary alignment and function. The extension tendon helps counteract spastic flexion about the metacarpophalangeal (MP) and interphalangeal (IP) joints, helps counteract adduction about the carpometacarpal (CMC) joint, and, in the active-thumb version, can be actuated to physically move the thumb into a wider grasp. The abduction tendon, always passive in our designs, provides support and rotates the thumb about the CMC into a more functional pose. The combined tendon network enables active opposition\slash reposition motions without requiring precise alignment of joint centers.

Unlike other 2-DoF thumb opposition approaches that require coordination from two thumb actuators \cite{rose2019}, or require the paretic limb to have little resistance to achieve their working principle \cite{butzer2020}, our method leverages tension between the two cables and the paretic thumb's bias towards flexion in order to create a lightweight mechanism that can be actuated with single motor. We leverage the stereotypical asymmetry of stroke impairment to reduce device complexity by strictly assisting extension while using body-powered movements for flexion. Our two-tendon routing also affords per-user, per-session customization to accommodate variance in spasticity. The main contributions of this paper can be summarized as follows:
\begin{itemize}
\item We introduce a novel cable mechanism leveraging tension between two cables to stabilize thumb opposition in the context of a robotic exoskeleton for upper extremity hemiparesis. The active version of the orthosis generates thumb motion using a single actuator, maintaining thumb stability throughout without requiring extraneous linkages. 
\item To the best of our knowledge, this is the first exoskeleton study to specifically evaluate and compare both active and passive thumb stabilization performance on an impaired population. Here we demonstrate a quantitative link between assisting thumb motion and observing functional improvement in grasp stability within an impaired population.
\end{itemize}
Our results show what the rehabilitative community has long intuited\textemdash stroke-impaired thumb opposition benefits from stabilization about a combination of axes, and improving the positioning of the thumb in opposition increases the ability to keep objects in-hand and complete more prolonged tasks.\\

\section{Mechanical Design}

The thumb mechanism presented here is used in conjunction with an existing hand orthosis, shown together in Fig. \ref{fig:myhand}, because our interest in studying effects of thumb assistance on hand function cannot be performed without also assisting the other digits. Design of the hand orthosis was previously described in detail in our previous work \cite{park2020,park2019,park2018}, but we include below a brief description of its most relevant aspects for ease of reference. We expand on our work with tendon-driven systems to develop an actuation system for the thumb, shown in Fig. \ref{fig:actuationdiagram}, that enables motion while stabilizing against individualized presentations of motor coupling and spasticity.

\subsection{Cable-driven Hand Orthosis}

The hand orthosis is a modular device that supports interchangeable mounting of individually-customized 3D-printed hand components to an aluminum splint that fixes the wrist at a neutral angle. Velcro straps around the hand and arm secure and locate the device. The forearm splint houses an actuated winch mechanism that connects to a tendon network routed through cable guides at the MP knuckles and anchored to 3D-printed fingertip components. Motorized retraction of the tendon network transmits finger-extension torques to the IP and MP joints, opening the hand.

The design of the knuckle-mounted cable guides and fingertip components preferentially transmits extension torques about the proximal IP joint while minimizing hyperextension about the MP. Dimensions of the fingertip components are sized for each subject's finger measurements; these plastic splints fixate the distal IP joint and are secured with Velcro straps. Foam padding and non-slip fabric straps enable these components to be tightly secured to the fingers while minimizing discomfort and migration.

Subjects press a button to switch between ``open" and ``closed" position setpoints, which are calibrated for each individual. With the active thumb condition, pressing the button commands both the hand-actuator and the thumb-actuator. Hand-actuator extension or retraction takes approximately 1.8 seconds, and the thumb actuator engages following a 1\textendash3 second delay after the hand actuator starts moving. We selected this delay based on spasticity level or user preference. Actuation uses integrated encoders for closed-loop PID motor control. For this prototype, motor drivers and power supply are located external to the device.

\subsection{Dual-cable Thumb Actuation}
Thumb motion is divided into actuated \mbox{flexion\slash extension} and passive \mbox{ab\slash adduction}, which together stabilize the thumb in opposition\slash reposition about the MP and CMC joints. This tendon network maintains the thumb in useful opposition at a range of pad-to-pad distances. As with the fingertip components, a rigid plastic splint fixates the distal joint to prevent hyperextension.

\begin{figure*}[t]
    \centering
    \vspace{.25cm}
    \includegraphics[width=0.99\textwidth]{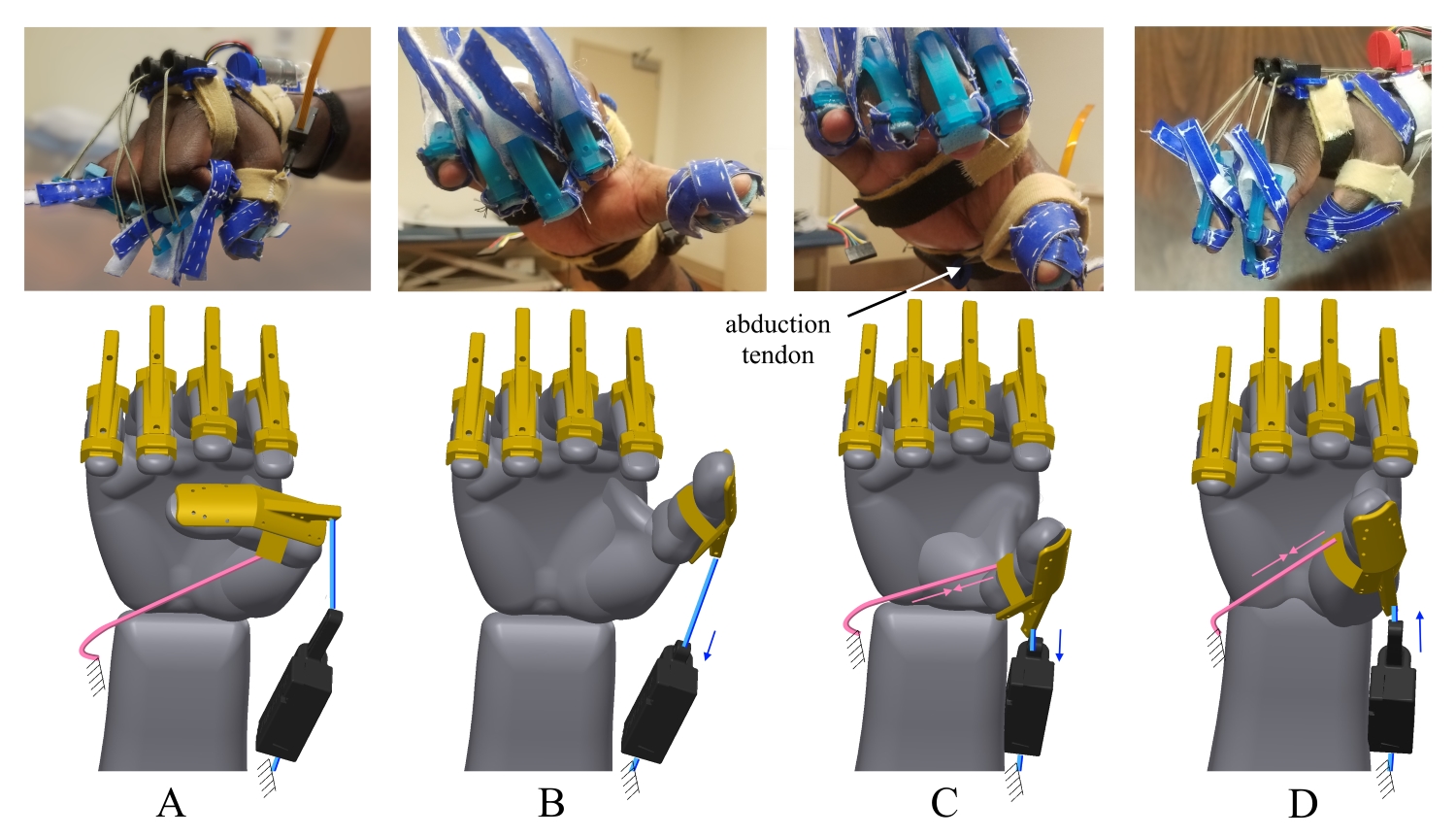}
    \caption{Dual-cable thumb opposition mechanism. \textit{Top}: stroke subject demonstrating hand configurations. \textit{Bottom}: working principle with passive abduction tendon (pink) and active extension tendon (blue). The default position for the paretic hand when both tendons are slack, i.e. disengaged, is a closed fist (Scenario A) as the stroke subject has voluntary flexion but not extension. Without the abduction constraint, active extension pulls the thumb away from the fingers (Scenario B). With a passive abduction tendon, the thumb is stabilized in opposition. The abduction tendon blocks the actuator from overextending the thumb; instead, the thumb rotates about the CMC joint (Scenario C). From this hand position, gradually releasing the actuator under body-powered flexion guides the thumb into a pad-to-pad grasp against the fingertips (Scenario D). When both tendons are completely slack, the thumb buckles against the fingers and returns to the original closed fist. The lengths of the extension and abduction cables are calibrated per subject at the start of the experiment.}
    \label{fig:actuationdiagram}
    \vspace{-10pt}
\end{figure*}

\begin{figure}[h!]
    \centering
    \includegraphics[width=.99\columnwidth]{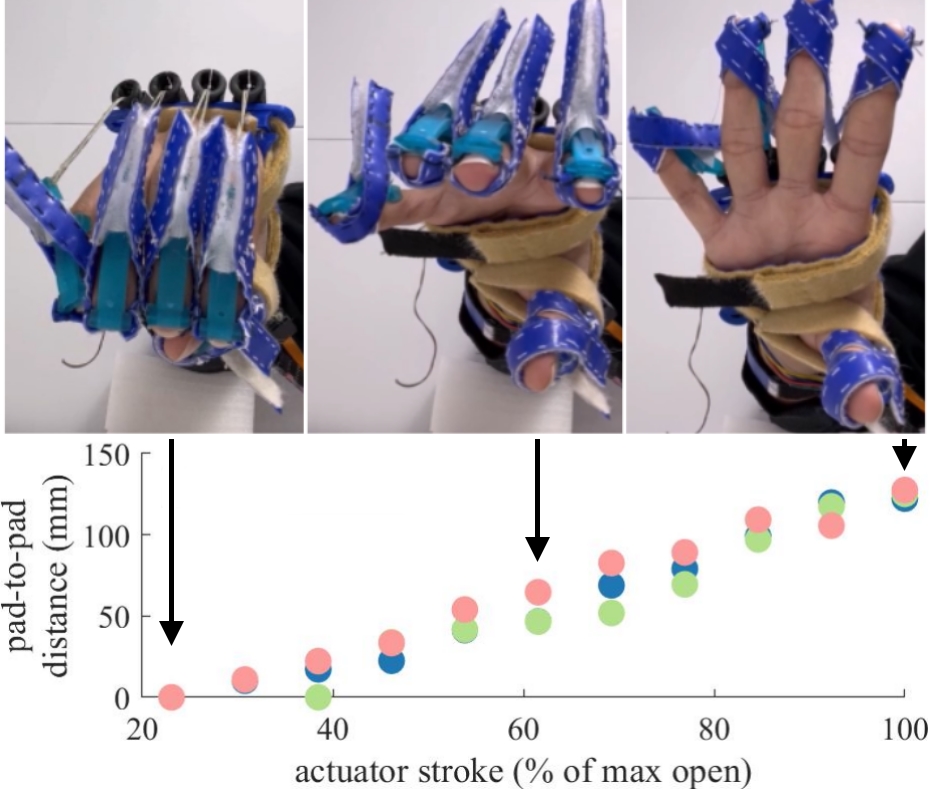}
    \caption{Healthy-subject demonstration to show index-thumb range of motion supported by the device. Top: photos left-to-right showing fully-closed, half-open, and fully-open hand poses. Bottom: measurements of pad-to-pad distances with three healthy subjects.}
    \label{fig:healthydemo}
    \vspace{-8pt}   
\end{figure}

The extension tendon is connected between the radial side of the forearm splint and the dorsal side of the thumb, crossing the splinted wrist. This tendon anchors to an outrigger feature on the dorsal side of the thumb splint, which enhances force transmission by increasing the moment arm about the MP joint. A linear actuator \mbox{(Actuonix-PQ12-P)} attaches to the rear of the forearm splint such that it is suspended at the radial side of the forearm, in-line with the thumb's path of extension. The actuator provides 20mm retraction when fully engaged---about 15$^\circ$ to 30$^\circ$ in extension depending on hand size and user preference. For the passive version of the device, the linear actuator is adjusted to user preference and is kept at a constant position for the duration of the experiment.

The abduction tendon is anchored to the ulnar side of the forearm splint, loosely located near the styloid process, and routes around the palmar side of the hand to secure the thumb's proximal link with a fabric hammock. Specifically, this passive tendon holds the thumb in abduction; pure extension without an abduction constraint pulls the thumb away from optimal opposition, as shown in \mbox{Fig. \ref{fig:actuationdiagram}B}.

The combination of active extension and passive abduction defines the trajectory space of thumb opposition. The degree of abduction constraint can be manually set by lengthening or shortening an adjustable grip-hitch knot securing the tendon to the thumbtip splint. Tension maintained between the two tendons throughout the thumb's motion counteracts reflexive flexion and adduction of the thumb when the digits are extended by the device. Within this study, we chose one general abduction-tendon length per subject for all of the experiments based on individual user preference, which was set to encourage index-thumb pinch grasps while still facilitating whole-hand power grasps (Fig. \ref{fig:actuationdiagram} C--D). Fig. \ref{fig:healthydemo} shows how this tendon network keeps the thumb oriented in opposition throughout the range of motion with a demonstration on healthy subjects.

\section{Experiments}

\begin{figure*}[t]
    \centering
    \vspace{.25cm}
    \includegraphics[width=0.99\textwidth]{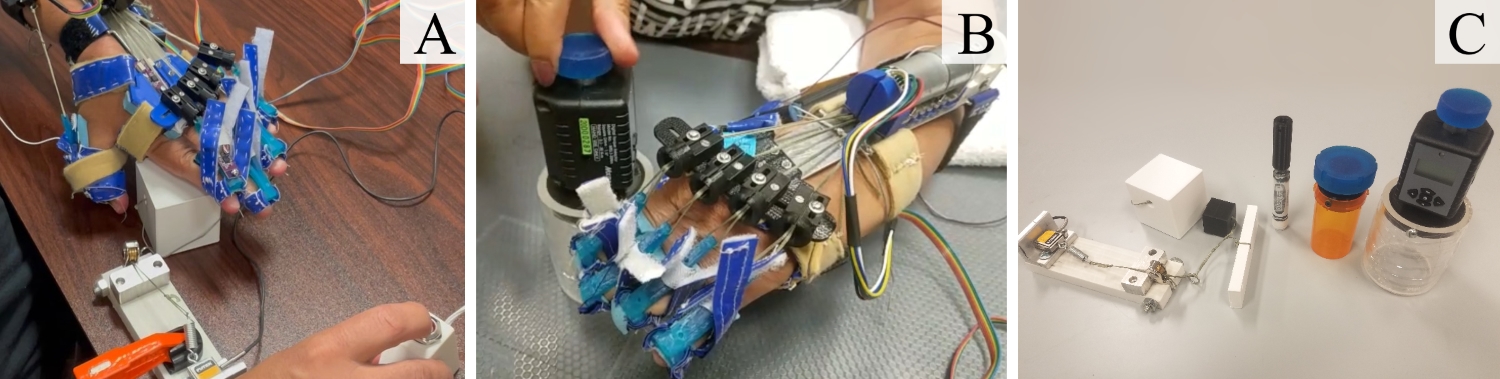}
    \caption{Photos of subjects completing unimanual pull tasks (A) and bimanual twist tasks (B), along with a photo of the objects used in this study (C).}
    \label{fig:experiment}
    \vspace{-5pt} 
\end{figure*}

Of the numerous functional grasp patterns afforded by hand anatomy, most healthy human subjects rely on pinch and power grasps that use the thumb extensively; stroke survivors lacking thumb function instead often rake the four fingers against an object to drag it along a surface, or tuck the thumb away against the backs of the fingers in order to squeeze an object between fingers and palm \cite{ga2017}. Our orthosis, like any exoskeleton with rigid elements spanning the hand, limits the ability to perform compensatory grasps that avoid thumb involvement. Our assessments of the thumb cable network have two objectives: 1) observe any improvement in orthosis performance due to thumb actuation compared to passive stabilization, and 2) determine whether stroke subjects gained functional improvement compared to not wearing a device at all.

Initial evaluation of the active thumb mechanism tested the users' ability to maintain stable grasps against disturbances for a range of object sizes. Primary assessments included: \mbox{1) grasping} and pulling a tethered object as hard as possible without letting the object slip out of the hand, and \mbox{2) stabilizing} cylindrical objects while exerting twisting torques. We conducted experiments over the course of two 90-minute sessions both with and without exoskeleton assistance of the impaired arm, and both with thumb actuation and the thumb splinted in gross opposition. The order of assistive conditions, and of tasks within each condition, were randomized to minimize effects of practice or fatigue. We allowed patients a few minutes at the start of each session to familiarize themselves with the open/close button, and allowed one practice trial with the object before each task. 

\subsection{Participants}
Five community-dwelling stroke survivors with chronic hemiparesis and limited upper-limb motor function volunteered to participate in the study. Eligible participants met the following inclusion criteria: (1) at least 18 years of age; (2) at least 6 months post-stroke; (3) muscle tone and spasticity scoring $\leq2$ on the Modified Ashworth Scale in digits, wrist, and elbow; (4) passive range of motion of digits and wrist within functional limits; (5) unable to extend fingers fully without assistance; (6) sufficient active flexion in digits, elbow, and shoulder to form a closed fist and lift the arm above table height; (7) intact cognition to provide informed consent and follow complex commands.

Subjects S1 and S2 had prior experience with early prototypes of the actuated-thumb exoskeleton, but not with the tasks described in this protocol. All stroke and healthy subjects provided informed consent to participate in this study in accordance with the protocol (IRB-AAAS8104) approved by the Columbia University Medical Center Institutional Review Board. Stroke participants were primarily recruited from a voluntary research registry of stroke survivors or referred from within NewYork-Presbyterian Hospital. All experiments with stroke subjects were performed under supervision of an occupational therapist.

\subsection{Unimanual Pull Task}

We evaluated subjects' ability to grasp an object, lift it clear from the table, then maintain the grasp while pulling as hard as possible against the object's tether without letting it slip out of the hand. The three 3D-printed objects were intended to be lightweight but in a range of sizes: a large cube (6cm sidelength), small cube (2.5cm sidelength), and thin rectangular prism (1.5cm width). The test apparatus consisted of the object tethered to a load cell (Futek LSB200-FSH00097, sampling rate 10Hz, resolution 0.01N), which was clamped to the table. An in-series pulley and stiff extension-spring mounted between object and load cell allowed subjects to freely pull without imposing jerk or off-axis loads on the sensor. We allowed for any type of grasp in which 1) the object was lifted clear of the table and 2) neither tether nor cable tension were involved in keeping the object in the hand (i.e. ``hook" grasps were disallowed). Invalid trials were repeated.

Subjects were instructed to pull the object against the tether, and keep resisting, as hard as they could without dropping the object; each trial concluded when the object slipped from grasp or after the subject maintained a constant arm position and accompanying tether tension for approximately five seconds. Participants repeated the task three times per object, per condition. Fig. \ref{fig:experiment}A shows a subject performing the unimanual-pull test with the large cube.

\subsection{Bimanual Twist Task}

We were also interested in evaluating device performance under conditions where the other hand could assist in object positioning and manipulation. We evaluated subjects' ability to ``open" cylindrical objects using a bimanual palmar grasp-torque test. In this experiment, the hand wearing the device would be expected to stabilize an object while the stronger hand performed the dexterous task. We used three simulated real-world objects: a water bottle (6.5cm diameter), pill bottle (4cm diameter), and marker (1.5cm diameter). These objects were attached to a digital torque meter (MXITA 0.3--30 Nm) along with a 3D-printed set of object-specific caps. 

Subjects were instructed to use their other hand to grab and place the object into the hand wearing the device as if they were to open the object. As soon as they had a stable grip, the stronger hand then twists the cap as hard as possible to achieve a peak-torque measurement; the cap does not experience angular displacement with this apparatus. During this experiment, the user gives open\slash close voice commands to a researcher who is operating the orthosis. Trials were videotaped to determine time elapsed between grabbing the object and beginning to exert torque on the cap. Participants repeated the task three times per object, per condition. Fig. \ref{fig:experiment}B shows a subject performing the bimanual-twist task with the water bottle.
\\
\section{Results and Discussion}

\begin{figure}[t]
\centering
\vspace{0.25cm}
\includegraphics[width=.9\columnwidth]{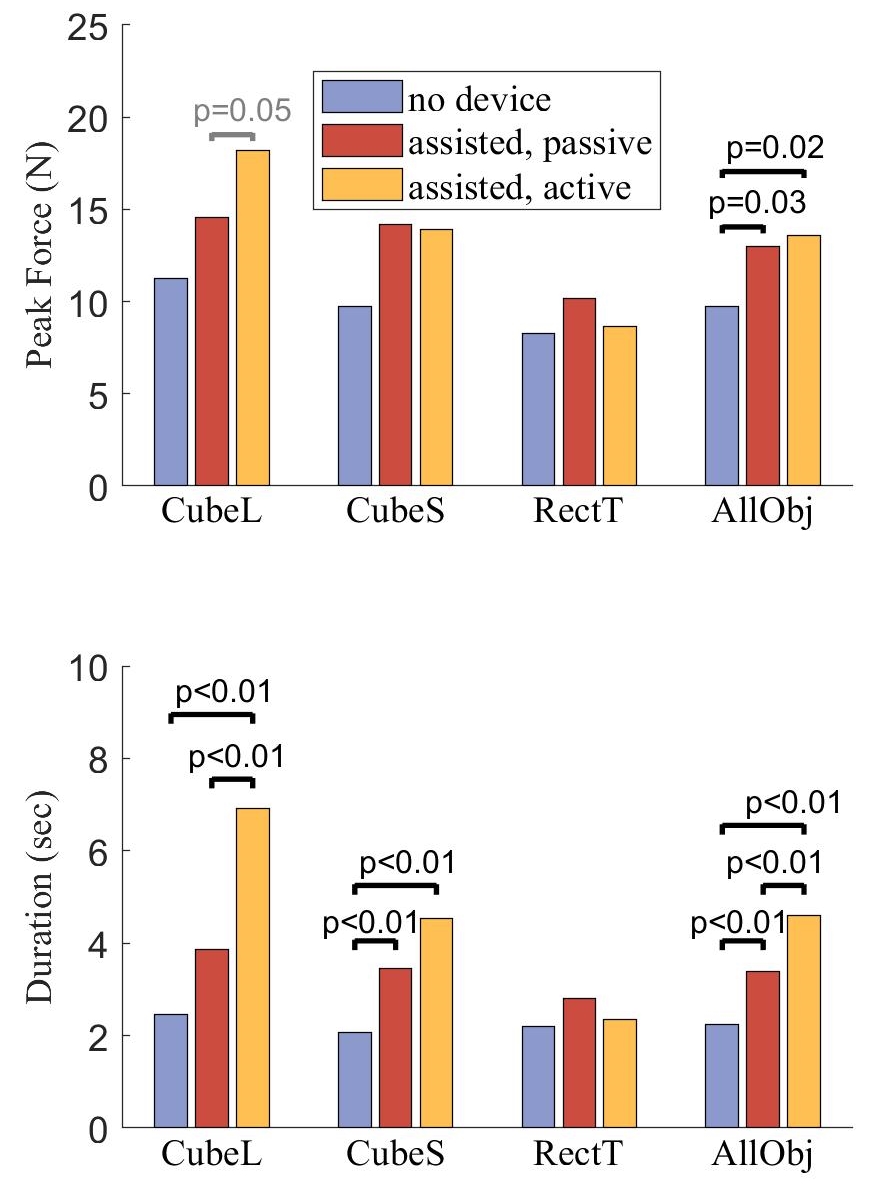}
\caption{Unimanual-pull peak force (Top) and duration (Bottom) results aggregated across all subjects. Displayed bars indicate median; p-values and bars indicate pairwise differences where significant or equaled $\alpha = 0.05$. Sample sizes were $n = 15$ per object, and $n = 45$ across all objects.}
\label{fig:unimanual}
\vspace{-10pt}
\end{figure}

\begin{figure}[t]
\centering
\vspace{0.25cm}
\includegraphics[width=.9\columnwidth]{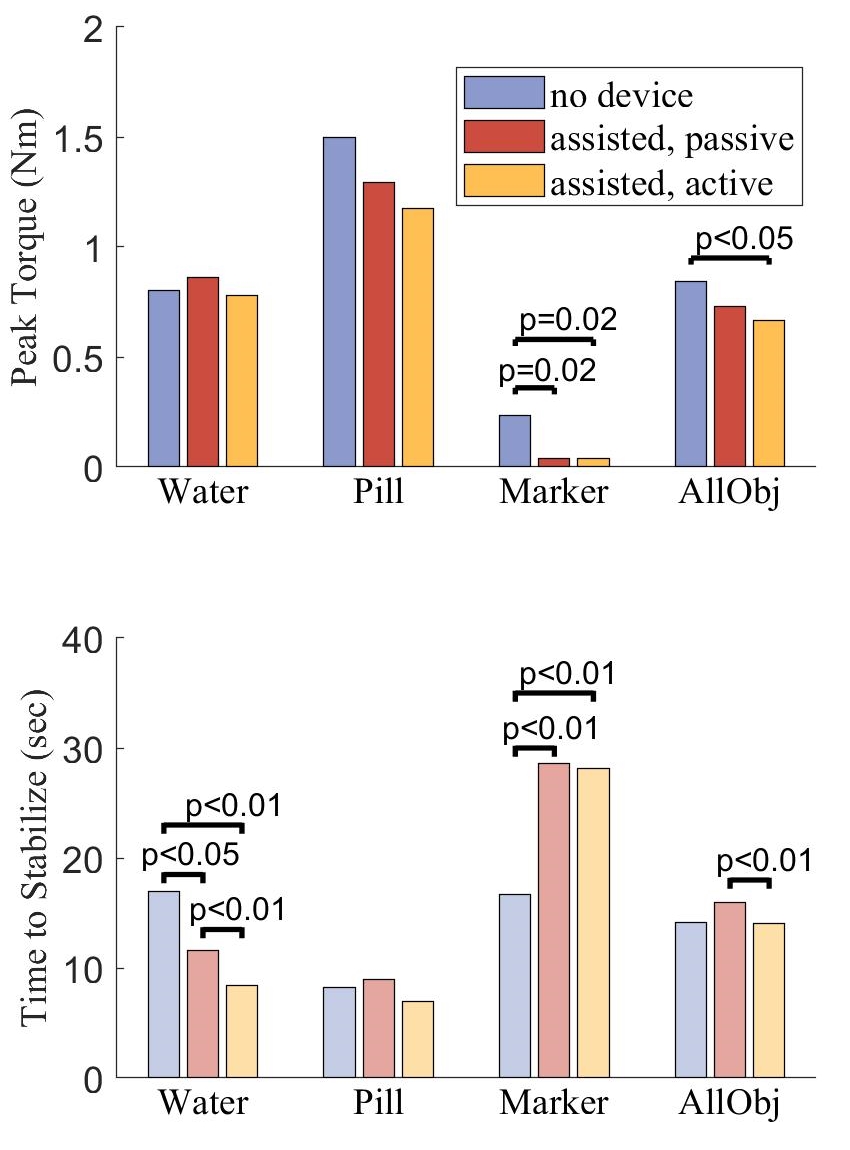}
\caption{Bimanual-twist peak torque (Top; $n = 15$ per object, $n = 45$ for all objects) and elapsed time to stabilize (Bottom; total $n = \lbrack 39, 41, 42\rbrack$ for \lbrack unassisted, passive, active\rbrack) results aggregated across all subjects. Displayed bars indicate median; p-values and bars indicate pairwise differences where significant. Time measurements could not be determined for $\lbrack 3, 1, 1\rbrack$ pill-bottle and $\lbrack 3, 3, 2\rbrack$ marker trials, which were excluded from analysis. Unlike the other performance metrics, elapsed-time evaluates whether assistance facilitated task speed (less time), thus is plotted with a different color-shading scheme.}
\label{fig:bimanual}
\vspace{-10pt}
\end{figure}

For each object and device condition, we report the aggregated result across all five subjects for unimanual-pull peak pulling force and time duration in Fig. \ref{fig:unimanual} and for bimanual-twist peak torque and time-elapsed-to-stabilize in Fig. \ref{fig:bimanual}. We include an overall summary of performance across all subjects and objects at the right of each figure. To examine the statistical significance in differences between active and passive thumb stabilization, and between no device assistance, we perform a dependent one-sided Wilcoxon signed rank test on the results aggregated across all subjects, using a hypothesis threshold $\alpha = 0.05$. We choose a non-parametric statistical test because we do not assume an underlying normal distribution, and choose a dependent test to reflect multiple samples drawn from each subject. Because our study population exhibits a range of dexterity impairments, we  specifically use medians to examine population trends in order to avoid skewing results towards large outliers in grip strength; for completeness, we also report the same unimanual-pull results aggregated across subjects as mean and standard deviation in Table \ref{tab:agg_data}.

Performance in the unimanual-pull task is evaluated along force and time components: we report both the peak pulling force and the time duration of the sustained pull before the object slips. We define a \textit{sustained pull} as the longest consecutive sequence of force readings above 1.0 N. The combination of measurements reflects how well the subject approached competing priorities of ``pull as hard as possible" and ``do not let the object slip out of the grasp." Trials for which subjects were unable to grasp the object were scored as zero for force and duration. In cases where subjects were able to hold on to the object indefinitely such that the object never slipped, each was assigned a duration of 10 seconds for comparison.

\begin{table*}[t]
    \vspace{0.25cm}
    \renewcommand{\arraystretch}{1.3}
    \caption{\textsc{Aggregated mean-average and standard deviation for peak force and duration of unimanual pull tests for 5 stroke subjects. Successful trials in which no slips occurred are reported as a 10-second duration. A summary for all-objects is included at the right. For each object, the best per-column result is shown in bold-text.}}
    \begin{minipage}[t]{\dimexpr.55\textwidth}
        \centering
        \textsc{\strut Peak Pull Force (N)}
        \begin{tabular}{c|ccc|c}
        \toprule
        \textbf{Condition}& \textbf{Large Cube} & \textbf{Small Cube} & \textbf{Thin Prism} & \textbf{All Objects}\\
        \midrule
        Active Thumb & \textbf{18.2 $\pm$ 11.0} & 13.9 $\pm$ 10.3 & 8.6 $\pm$ 12.9 & \textbf{13.6 $\pm$ 11.9}\\
        Passive Thumb & 14.6 $\pm$ 11.9 & \textbf{14.2 $\pm$ 12.1} & \textbf{10.1 $\pm$ 12.9} & 13.0 $\pm$ 12.2\\
        No Device & 11.3 $\pm$ 17.7 & 9.8 $\pm$ 17.6 & 8.3 $\pm$ 14.4 & 9.8 $\pm$ 16.3\\
        \bottomrule
        \end{tabular}
    \end{minipage}\hfill
    \begin{minipage}[t]{\dimexpr.45\textwidth}
        \centering
        \textsc{\strut Duration of Sustained Pull (sec)}
        \begin{tabular}{ccc|c}
        \toprule
        \textbf{Large Cube} & \textbf{Small Cube} & \textbf{Thin Prism} & \textbf{All Objects}\\
        \midrule
        \textbf{6.9 $\pm$ 3.9} & \textbf{4.5 $\pm$ 3.9} & 2.3 $\pm$ 4.1 & \textbf{4.6 $\pm$ 4.3}\\
        3.9 $\pm$ 3.7 & 3.5 $\pm$ 3.8 & \textbf{2.8 $\pm$ 4.5} & 3.4 $\pm$ 3.9\\
        2.5 $\pm$ 4.1 & 2.1 $\pm$ 4.1 & 2.2 $\pm$ 4.1 & 2.2 $\pm$ 4.0\\
        \bottomrule
        \end{tabular}
    \end{minipage}
    \label{tab:agg_data}
\end{table*}
 
As shown in Fig. \ref{fig:unimanual}, active-thumb assistance outperformed passive-thumb assistance in overall force generation and time duration. Both versions of assistance were better than no-device. Differences in relative performance between the assistance conditions were more apparent for larger objects. To our surprise, we did not find either active- or passive-thumb assistance to have a significant per-object force improvement over the unassisted condition. However, active-thumb assistance was significantly better than both passive and unassisted conditions in keeping a grip on most objects for a longer duration. Passive-thumb also outperformed the unassisted condition in duration for all but the thin rectangular prism. 

Our device specifically assists digit extension, not flexion or shoulder strength; thus, pull force and time results together reflect a measurement of task success rate. If an object is held in a stable grasp, we expect subjects to be able to repeatedly pull against the object for an indefinite amount of time without dropping it. If subjects were not confident in their grips on the object, subjects might opt for pulling with less strength in order to optimize duration. Alternatively, subjects might try to maximize force if they believe the object is certain to slip. Both force and time results are reduced in cases where subjects were unable to consistently grip and pull an object with any meaningful force, such as with the thin rectangular prism.

The bimanual torque experiment was designed to evaluate whether the bulk of the device hindered gross grasping, since the other hand places the object into the hand wearing the device and performs the more-dexterous job of twisting the cap. Again, we consider both torque and time components: we expect assisting hand-opening to make it easier to place larger objects in the hand, so a successful result should require less time to achieve similar torques. These results are shown in Fig. \ref{fig:bimanual}. Here, the role of the assisted hand is simply to stabilize the object and resist torque. Comparing active-thumb assistance to the passive version, we found that the added complexity did not significantly affect torque generation but did reduce time to position the object in the hand to begin grasping. Contrary to expectations, only subject S4 was able to exert torques on the marker in any device condition, and this subject achieved similar torque measurements with and without the device. Overall, subjects could not achieve as high torques when wearing the device as they could without it, and active thumb assistance was neither significantly faster or slower than the unassisted condition. 

In both experiments, we expected device assistance to improve subjects' performance with larger objects but have less clear benefit with smaller objects. Although our results support this trend, our study was limited in that, because we do not assist finger flexion, our smallest objects could either be grasped in all cases or none depending on the individual. Our choice of objects reflected incorrect assumptions that the passive version of the device would fail outside of a narrow range of sizes, and that the active version of the device could facilitate pad-to-pad grasps for all subjects. Further exploration with a larger set of subjects and objects is needed to enhance our understanding of functional thumb assistance.

Our overall results find active-extension thumb assistance to improve consistency in generating and maintaining grasp forces for a range of object sizes. In particular, thumb actuation has a comparative advantage over passive splinting when precise positioning and preshaping of the hand is required for manipulation. When comparing between unimanual and bimanual performance, we see that thumb actuation has a much greater effect in reducing likelihood of dropping an object when the impaired hand must reach for it; when the hand simply needs to open and accept an object as part of a bimanual task, differences in force generation become negligible. Looking at the per-object aggregation, we observe that assisted-opening of the hand can help with grasping larger objects without having to sacrifice performance for smaller objects.


\section{Conclusions and Future Work}

In this paper, we present a two-tendon actuation method to assist thumb opposition\slash reposition for a hand orthosis. The proposed design uses one cable routed around the palmar side of the hand to secure the thumb in abduction and a second cable attached to the dorsal side of the thumb to enable extension. This design allows for an active stabilization mechanism: the flexion-extension cable is connected to a linear actuator to provide thumb motion in one degree of freedom, while the abduction cable remains passive providing static support. Our thumb stabilization design is integrated into a robotic orthosis that also provides assistance for extension of the other four digits, thus enabling functional tasks. 

We demonstrate functional efficacy of our exoskeleton in actively assisting thumb extension to improve grasp stability, for which we examined both the magnitude of resistive force when pulling against an object and the amount of time the user was able to maintain adequate contact forces to keep the object from slipping.  We performed these evaluations with stroke survivors having spasticity and limited hand function in order to study real-life applicability. We uniquely evaluated our actuation approach not only against a baseline condition where no device is worn, but also against the configuration that provides only passive thumb assistance by fixing the length of both tendons. 

Our results show that active thumb stabilization allows users to maintain grasp forces for longer duration than its passive counterpart in the case of unimanual tasks, where the impaired hand must reach for and grasp the target object. Furthermore, an orthosis using the proposed thumb stabilization method (either active or passive) enables greater grasp force magnitude and duration compared to a no-device baseline. However, these advantages were not present for bi-manual tasks, where the unimpaired hand can be used to assist grasp formation.

In future work, we would like to further explore the temporal aspects of grasp stability. Our study challenged participants to exert high arm-flexor effort to pull and twist objects, triggering spastic motor synergies and increased muscle tone---a modified set of tasks that tracked how quickly the hand could release objects after successful task completion would complement our work on finger-thumb extension. We hope our work inspires others to also consider grasp-durations when conducting device evaluations; this would bring the field more in-line with the rehabilitative needs of this stroke population.

Finally, we continue working towards our main goal: developing a wearable orthosis that can assist stroke survivors in everyday activities outside of a structured research environment. Numerous challenges must still be overcome to realize this vision, both in effectiveness (improved function over a wider range of metrics) and usability (more intuitive and streamlined design). As different tasks may benefit from different device configurations, effective devices must be adaptable to meet all of the functional demands of the hand. The intuitive and easily customizeable nature of tendon-based actuation lends itself well to encouraging patients' engagement and creativity with the device, and our participants suggested possible passive-active tendon combinations to aid additional tasks. We believe that further research on wearable robots, when conducted in partnership with stroke survivors, can achieve highly capable devices with the potential to meet users' diverse needs in assisting and encouraging use of the impaired limb after stroke.


\end{document}